\documentclass[letterpaper]{article} 
\usepackage{aaai2026}  
\usepackage{times}  
\usepackage{helvet}  
\usepackage{courier}  
\usepackage[hyphens]{url}  
\usepackage{graphicx} 
\usepackage{natbib}  
\usepackage{caption} 
\usepackage{algorithm}
\usepackage{algorithmic}

\usepackage{amsmath}
\usepackage{booktabs}
\usepackage{mathtools}
\usepackage{amssymb}
\usepackage{amsthm}
\usepackage{cleveref}
\usepackage{bm}
\usepackage{dsfont}
\usepackage{pifont}
\newtheorem{theorem}{Theorem}

\newtheorem{proposition}{Proposition}
\newtheorem{assumption}{Assumption}

\usepackage{newfloat}
\usepackage{listings}
\DeclareCaptionStyle{ruled}{labelfont=normalfont,labelsep=colon,strut=off} 
\floatstyle{ruled}
\newfloat{listing}{tb}{lst}{}
\floatname{listing}{Listing}
\title{GenePheno: Interpretable Gene Knockout-Induced Phenotype Abnormality Prediction from Gene Sequences}
\author {
    Jingquan Yan\textsuperscript{\rm 1}\equalcontrib,
    Yuwei Miao\textsuperscript{\rm 1}\equalcontrib,
    Lei Yu\textsuperscript{\rm 2},
    Yuzhi Guo\textsuperscript{\rm 1},
    Xue Xiao\textsuperscript{\rm 2},
    Lin Xu\textsuperscript{\rm 2},
    Junzhou Huang\textsuperscript{\rm 1}\thanks{Corresponding author.}
}
\affiliations {
    \textsuperscript{\rm 1}Department of Computer Science and Engineering, University of Texas at Arlington\\
    \textsuperscript{\rm 2}Quantitative Biomedical Research Center, School of Public Health, University of Texas Southwestern Medical Center\\
    \{jingquan.yan, yuwei.miao, yuzhi.guo, jzhuang\}@uta.edu, \{lei.yu, xue.xiao, lin.xu\}@utsouthwestern.edu
}

\begin{document}

\maketitle

\begin{abstract}
Exploring how genetic sequences shape phenotypes is a fundamental challenge in biology and a key step toward scalable, hypothesis-driven experimentation. The task is complicated by the large modality gap between sequences and phenotypes, as well as the pleiotropic nature of gene–phenotype relationships. Existing sequence-based efforts focus on the degree to which variants of specific genes alter a limited set of phenotypes, while general gene knockout-induced phenotype abnormality prediction methods heavily rely on curated genetic information as inputs, which limits scalability and generalizability. As a result, the task of broadly predicting the presence of multiple phenotype abnormalities under gene knockout directly from gene sequences remains underexplored. We introduce GenePheno, the first interpretable multi-label prediction framework that predicts knockout-induced phenotypic abnormalities from gene sequences. GenePheno employs a contrastive multi-label learning objective that captures inter-phenotype correlations, complemented by an exclusive regularization that enforces biological consistency. It further incorporates a gene function bottleneck layer, offering human-interpretable concepts that reflect functional mechanisms behind phenotype formation. To support progress in this area, we curate four datasets with canonical gene sequences as input and multi-label phenotypic abnormalities induced by gene knockouts as targets. Across these datasets, GenePheno achieves state-of-the-art gene-centric $F_{\text{max}}$ and phenotype-centric AUC, and case studies demonstrate its ability to reveal gene functional mechanisms.
\end{abstract}

\section{Introduction}
\label{sec:introduction}
Understanding the relationship between genetic information and phenotypic outcomes has long been a fundamental goal in biology. Genetic information, encoded in gene sequences, influences phenotypes through complex yet broadly consistent biological mechanisms across individuals~\citep{feuermann2025compendium}. Predicting the phenotypic outcome from gene information is fundamental and essential to therapeutic discovery~\citep{tang2024rnai}, functional genomics~\citep{ zheng2024leveraging}, and systems biology~\citep{whiting2024phenotypic,islam2025learning}. Existing approaches generally follow two directions. Variant effect prediction methods typically take gene sequences as inputs to estimate how specific genetic variants change the magnitude or degree of a limited set of phenotypes. Other methods for predicting large-scale phenotypic abnormalities at the gene or protein level rely heavily on labor-intensive curated information, such as protein–protein or gene–gene interaction networks, which limits their applicability to newly discovered or poorly annotated genes. Moreover, both types of methods offer limited insight into the intermediate functional mechanisms that connect genetic information to phenotypic outcomes. These limitations highlight a critical gap: the need for interpretable methods that can predict large-scale phenotypic abnormalities directly from gene sequences, thereby improving applicability to under-annotated genes and enhancing interpretability of the underlying biological mechanisms.

\begin{figure*}[t]
    \centering
\includegraphics[width=1\textwidth]{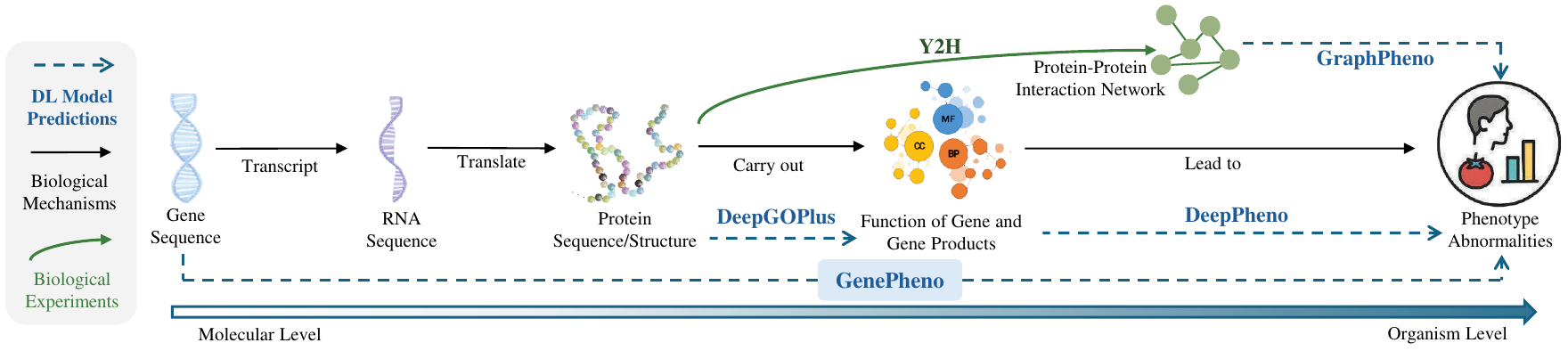}  
    \caption{Overview of key biological modalities linking molecular-level DNA sequences to organism-level phenotypic traits and representative methods modeling different stages. Our method, GenePheno, bridges the modality gap in an end-to-end manner.}
    \label{fig:biology_pipeline}
\end{figure*}

To address this gap, it is important to understand the biological flow of information from gene sequences to phenotypes. Research into the relationship between genetic information and phenotypic traits generally follows the central dogma of molecular biology.~\Cref{fig:biology_pipeline} outlines the key biological modalities involved in passing genetic information from molecular-level gene sequences to organism-level phenotypes. At the molecular level, genetic information encoded in the linear sequence of DNA is transcribed into RNA and translated into proteins, whose amino acid composition and three-dimensional conformation determine their biochemical properties and functional roles within the cell~\citep{crick1970central}. The functions of genes and their products are systematically organized in a directed acyclic graph (DAG) known as the Gene Ontology (GO)~\citep{gene2019gene}. Predicting GO terms from protein sequence or structure has long been an active area of research~\citep{kulmanov2020deepgoplus, yuan2023fast, liu2024interlabelgo+, Gligorijevic2019}. Further attempts are made to predict large-scale knockout-induced phenotype abnormalities directly from the GO terms annotated to the corresponding protein~\citep{kulmanov2020deeppheno}. Besides general GO functions, a key functional property of proteins is their capacity to interact with one another. Protein–protein interaction (PPI) networks, derived from assays such as yeast two-hybrid (Y2H), offer a curated modality of genetic information. Although PPI networks are widely used in prediction of large-scale phenotype abnormalities, their utility is limited to proteins with experimentally validated interactions~\citep{ bi2023sslpheno, liu2022integration}.

Phenotypic abnormalities are systematically organized in DAGs such as the Human Phenotype Ontology (HPO) and Mammalian Phenotype Ontology (MPO), where nodes represent phenotype abnormalities and edges encode subsumption relationships~\citep{gargano2024mode, smith2009mammalian}. Due to pleiotropy and overlapped functional mechanisms, phenotypes often show strong dependencies and correlations~\citep{mackay2024pleiotropy}. Such dependencies are evident in genes whose functions influence multiple phenotypes, such as those involved in pigmentation, which are also associated with hearing or vision abnormalities~\cite{reissmann2013pleiotropic}. However, most existing studies formulate the multi-label phenotype abnormality prediction as multiple binary classification tasks, neglecting correlations between phenotypes~\citep{bi2023sslpheno, liu2022integration}. These correlations include logical constraints, where certain phenotype abnormalities are mutually exclusive. For example, in HPO, abnormal muscle tone includes both hypotonia (reduced muscle tone) and hypertonia (increased muscle tone), which are semantically incompatible and should not co-occur. Existing prediction methods overlook these constraints, potentially producing logically inconsistent results.

To address these challenges, we propose GenePheno, the first interpretable multi-label prediction framework that predicts knockout-induced phenotypic abnormalities from gene sequences. GenePheno integrates GO functions at both fine and coarse granularity. Fine-grained GO terms, representing specific biological functions, are input alongside the gene sequence and fused via cross-attention. Coarse-grained GO categories serve as supervision targets at the bottleneck layer, providing human-interpretable concepts that reflect general mechanisms of phenotype formation. We further design a contrastive multi-label objective for phenotype abnormality prediction, capturing inter-phenotype correlations, and apply an exclusivity regularization to enforce biological logic consistency. By explicitly modeling the gene sequences, GO functions, and phenotype abnormalities, our work takes a significant step toward understanding how genetic information encodes large-scale observable traits. 

Our key contributions are summarized as follows: (1) To the best of our knowledge, we are the first to formally formulate the deep learning task of predicting gene knockout-induced phenotype abnormality directly from gene sequences. (2) We curated four benchmark datasets with a stratified split to support future research in this area. (3) We propose GenePheno, the first interpretable end-to-end framework that maps gene sequences to multi-label phenotype abnormalities, guided by a biologically motivated objective that captures phenotype correlations and enforces phenotype exclusivity. (4) GenePheno offers an interpretable architecture with a functional bottleneck layer. We perform case studies showing that the resulting interpretations are consistent with the biological understanding of phenotypes. (5) We conduct comprehensive experiments showing that GenePheno achieves strong performance in both gene-centric $F_{\max}$ and phenotype-centric AUC, with ablation studies validating the contributions of each component.

\section{Related Works}
\paragraph{Phenotype Prediction from Curated Genetic Information}
Previous studies commonly formulate the multi-label phenotype prediction as multiple binary classification task and their input mostly rely on curated genetic information such as gene expression profiles, GO annotations, or PPI networks. For example, DeepPheno~\citep{kulmanov2020deeppheno} predicts HPO terms using protein GO functions and expression data from 53 tissues. HPOFiller~\citep{liu2021hpofiller} constructs a bipartite graph connecting proteins and HPO terms, using GCN-based embeddings to infer associations. HPODNets~\citep{liu2022hpodnets} processes three different PPI networks with parallel GCN pipelines and fuses the outputs for HPO prediction. GraphPheno~\citep{liu2022integration} encodes protein sequences using Composition–Transition (CT) features as node inputs for a GCN over the PPI network. SSLPheno~\citep{bi2023sslpheno} integrates PPI and GO information to construct an attributed gene network and employs self-supervised contrastive learning to learn representations for predicting HPO terms via a downstream DNN classifier. Despite their contributions, these approaches face two fundamental limitations: they require pre-processed modalities, restricting their application to well-studied genes, and they fail to capture the biological interdependence and logical constraints among phenotypes by treating each as an isolated prediction task.

\paragraph{Sequence-based Genetic Property Prediction}
Recent efforts on sequence-based modeling have primarily focused on predicting GO functions from amino acid sequences. While many studies leverage curated genetic modalities such as PPI networks or 3D structural data to enhance performance, here we highlight approaches that rely primarily on protein sequence inputs. DeepGOPlus~\citep{kulmanov2020deepgoplus} uses one-hot encoded protein sequences followed by convolution and max-pooling layers to predict GO terms. SPROF-GO~\citep{yuan2023fast} encodes sequences with ProtT5~\citep{elnaggar2021prottrans}, pools residue embeddings via gated attention, and refines predictions using homology-based diffusion from DIAMOND~\citep{buchfink2015fast}. InterLabelGO~\citep{liu2024interlabelgo+} employs a rank-based loss and uses mean-pooled ESM-2 embeddings from the last three layers, followed by DIAMOND-based alignment and k-nearest-neighbor retrieval for GO term prediction. 

Existing gene sequence-based methods such as DeepSEA~\cite{zhou2015predicting}, Basset~\cite{kelley2016basset}, ExPecto~\cite{zhou2018deep}, Enformer~\cite{avsec2021effective}, AlphaMissense~\cite{cheng2023accurate}, EVE~\cite{frazer2021disease}, GENERator~\cite{wu2025generator}, and Evo2~\cite{brixi2025genome} emphasize on predicting the effect of local allelic variants by comparing short genomic windows under reference and alternate alleles. These methods differ fundamentally from our task, which involves modeling full-gene knockouts to predict the presence or absence of organism-level phenotypic abnormalities. Nonetheless, their advances in long-range DNA encoding provide valuable foundations for our encoder.

\section{Problem Formulation}
Let $\mathcal{G} = \{g_1, g_2, \ldots, g_N\}$ be the set of $N$ genes, and let the training set be $\mathcal{D}=\{(X_i,\mathbf{y}_i)\}_{i=1}^N$, where each $X_i\in\mathcal{X}$ denotes the input features for gene $g_i$ and $\mathbf{y}_i\in\{0,1\}^C$ is the binary label vector for $C$ phenotypes. We aim to learn a mapping $f_\psi:\mathcal{X}\to[0,1]^C$ by solving $\min_\psi \frac1N\sum_{i=1}^N \mathcal{L}\bigl(\mathbf{y}_i, f_\psi(X_i)\bigr)$ where $\mathcal{L}(\cdot,\cdot)$ is an appropriate binary or multi-label classification loss function. Below, we detail the construction of input $X_i$ for two representative methods. 

\textbf{PPI-based Method}
We encode protein-protein interaction as an undirected graph $G=(V,E)$ with adjacency matrix $A\in\{0,1\}^{|V|\times|V|}$, where $V = \{v_1, v_2, \ldots, v_N\}$ are the proteins corresponds to genes in $\mathcal{G}$. Let $H^{(0)}\in\mathbb{R}^{|V|\times d}$ denote the matrix of initial node features. A graph neural network computes node embeddings $H = \mathrm{GNN}_\phi(A, H^{(0)}) \in \mathbb{R}^{|V|\times d}$ and we write $ \mathbf{x}_i = H_{(v_i,:)}$ for the embedding of node $v_i$. Then we predict multi-label outputs with $\mathbf{\hat{y}}_i = f_\psi(\mathbf{x}_i) \in [0,1]^C$.

\textbf{Gene Sequence-based Method}
For each gene $g_i \in \mathcal{G}$, let $\mathrm{seq}_i$ denote its gene sequence. We first extract a token-level embedding via a pretrained sequence model $f_\theta$ to get the token-level embedding $\mathbf{e}_i = f_\theta(\mathrm{seq}_i) \in \mathbb{R}^{\ell\times d}$ where $\ell$ is the number of tokens. We then aggregate these token embeddings into a sequence-level embedding $\mathbf{x}_i=\operatorname{AGG}(\{\mathbf{e}_i\})$ and compute the multi-label output $\mathbf{\hat{y}}_i = f_\psi(\mathbf{x}_i) \in [0,1]^C$.
\begin{figure*}[t]
    \centering
\includegraphics[width=1\textwidth]{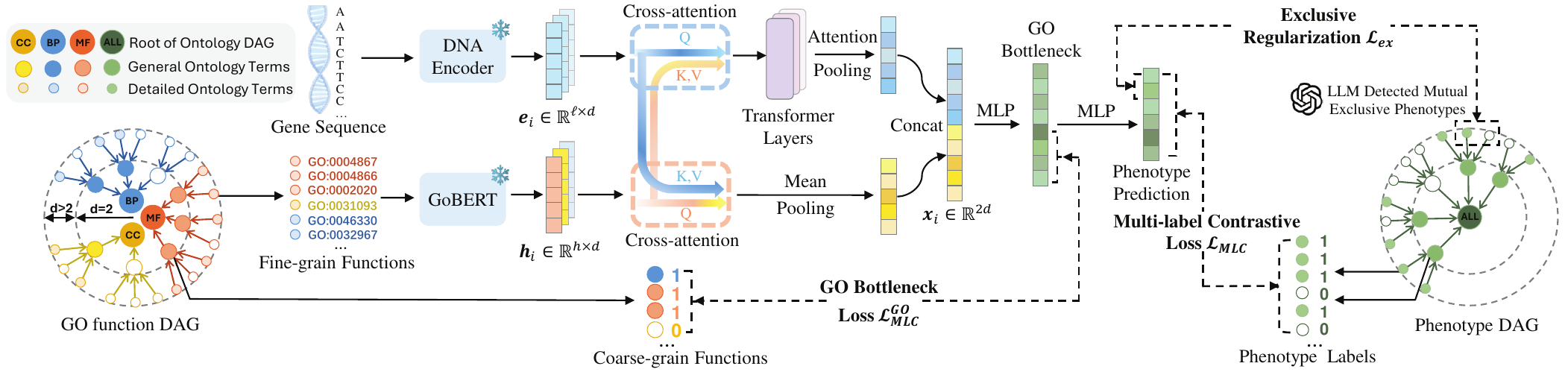}  
    \caption{Overview of our learning framework. The GO function DAG comprises three subgraphs, each with its own root, while the phenotype ontology DAG consists of a single graph with one root. In both structures, node depth $d$ denotes the shortest path to the root, with deeper nodes representing more specific functions or phenotypes. We utilize GO functions at dual granularity: fine-grain inputs ($d>2$) for detailed functional information, and coarse-grain bottleneck supervision ($d=2$) for general mechanisms. Target phenotypes span both general and specific categories.} 
    \label{fig:our_model}
    \vspace{-12pt}
\end{figure*}

\section{Method}

In this section, we present the framework of our proposed GenePheno model. We systematically address the aforementioned challenges of label correlation, phenotype exclusivity, and interpretability requirements in the following subsections. Finally, we outline the comprehensive architecture of our learning framework and formulate its overall learning objective.

\subsection{Multi-label Prediction}
\label{sec:multi-label prediction}
While phenotype prediction is typically formulated as a multi-label classification (MLC) task using binary cross-entropy (BCE) loss, this approach suffers from two limitations. First, BCE loss assumes label independence, failing to capture the inherent co-occurrence patterns in phenotypes that often exist in ontology hierarchies (e.g., ``Astigmatism'' is an ``Abnormality of the eye''). Second, BCE yields higher-order exponential terms involving positive and negative classes~\citep{su2022zlpr}, where predominating negative classes will exacerbate the class imbalance. To address these limitations, we derive our supervised contrastive objective from InfoNCE~\citep{oord2018representation}, traditionally used in unsupervised settings, to: (i) ties together all positive labels, which implicitly models label dependencies, and (ii) pools scores across positives and negatives globally.

To derive the contrastive MLC loss, we extend the InfoNCE loss to accommodate multiple positive and negative labels. Let $s_i(x)$ represent the classification logit for phenotype label $i$, $\Omega_{+}$ the set of positive phenotype labels, and $\Omega_{-}$ the set of negative phenotype labels. Our contrastive MLC loss consists of two components:
\begin{align}
\label{eq:multi-label_loss}
\mathcal{L}_{\mathrm{NCE}}^+
&= \sum_{i\in\Omega_{+}}
\log\!\big(e^{\frac{s_i(x)}{\tau}}+\sum_{j\in \Omega_{-}}e^{-\frac{s_j(x)}{\tau}}\bigr)
-\frac{s_i(x)}{\tau} \\
\mathcal{L}_{\mathrm{NCE}}^-
&= \!-\!\!\sum_{j\in\Omega_{-}}
\log\bigl(e^{-\frac{s_j(x)}{\tau}}+\sum_{i\in \Omega_{+}}e^{\frac{s_i(x)}{\tau}}\bigr)
+\frac{s_j(x)}{\tau}
\end{align}
where $\tau$ is a temperature hyperparameter. This formulation encourages clustering of logits within the positive and negative label sets while maximizing the distance between two clusters. Specifically, $\mathcal{L}_{\mathrm{NCE}}^+$ pulls positive labels together while pushing them away from negatives, and $\mathcal{L}_{\mathrm{NCE}}^-$ ensures negative labels remain clustered and distinct from positives. Our final contrastive MLC learning objective is the sum of the positive and negative components:
\begin{align}
    \mathcal{L}_\mathrm{MLC} = \mathcal{L}_{\mathrm{NCE}}^+ + \mathcal{L}_{\mathrm{NCE}}^-
\end{align}
The complete derivation from InfoNCE is detailed in \Cref{app:loss_derivation}. Our contrastive loss effectively captures label dependencies through the explicit separation of positive and negative label clusters, while simultaneously avoiding high-order exponential terms that exacerbate class imbalance. Notably, the ZLPR loss~\citep{su2022zlpr} used in InterLabelGO can be derived from our contrastive loss when $\tau = 1$ with the denominator shift $s(x)/\tau$ omitted. This equivalence confirms our loss inherits ZLPR's established properties of capacity to capture label correlations and robustness to label imbalance (see \Cref{app:zlpr_connection}).

\subsection{Phenotype Exclusivity Regularization}
\label{sec:exclusive}
Despite effectively capturing label co-occurrence and mitigating class imbalance, the contrastive MLC loss fails to incorporate semantic exclusivity in phenotype ontologies. Many phenotype abnormalities are inherently mutually exclusive—an organism cannot simultaneously 
exhibit both ``hypotonia'' and ``hypertonia'' traits. These exclusivity constraints represent critical biological knowledge, yet the standard contrastive MLC loss lacks mechanisms to encode these semantic priors, potentially generating biologically implausible predictions. This issue is exacerbated by the sparse nature of phenotype annotations, which impedes the model to learn these constraints from data alone. To address this limitation, we propose a 2-step complementary approach: (1) a large language model (LLM) pipeline for automated identification of mutually exclusive phenotype pairs that leverages ontological structure, and (2) a soft exclusive regularization term incorporated into our training objective to enforce biological constraints.

Let $\mathcal{O} = (P, E)$ be a phenotype ontology where $P$ represents the phenotype set and $E \subseteq P \times P$ denotes phenotype relationships. For phenotype $p_i \in P$, we define $D(p_i)$ as its set of direct descendants. Using an LLM with a structured query prompt (more details in~\Cref{app:llm_prompt}), we identify $\mathcal{E}_i = \{(q_k, q_j) \in D(p_i) \times D(p_i) \mid p_k \text{ and } p_j \text{ are mutually exclusive}\}$, the set of mutually exclusive phenotype pairs among direct descendants of $p_i$. We then compute this set for all non-leaf phenotypes in the ontology graph and obtain the dataset-level exclusivity set $\mathcal{E} = \mathcal{E}_1\bigcup\dots\bigcup \mathcal{E}_n$.

For any pair of exclusive phenotype in $\mathcal{E}$, we impose a soft regularization on the exclusive pairs' prediction logits with Softplus. The exclusive regularization  $\mathcal{L}_\text{ex}$ can be written as:
\begin{equation}
\begin{aligned}
    \mathcal{L}_\mathrm{ex} =\frac{1}{N}\sum_{n=1}^N \sum_{(i,j)\in\mathcal{E}}
\log\!\bigl(1+e^{\,s_i(\mathbf{x}_n)+s_j(\mathbf{x}_n)}\bigr)
\end{aligned}
\end{equation}
where $\mathcal{E}$ denotes a set of mutually‑exclusive label pairs and $s_k\in\mathbb{R}$ is the logit for label~$k$. We demonstrate the effectiveness of this loss using stationary analysis of gradient as follows.

\begin{proposition}[Stationary Analysis of Exclusive Regularization]\label{prop:exclusive_stationary}
Let $\mathcal{L}$ be the contrastive MLC loss with our proposed soft exclusive regularization weighted by any $\lambda>0$, namely $\mathcal{L}\;=\;\mathcal{L}_{\mathrm{MLC}}\;+\;\lambda \mathcal{L}_\mathrm{ex}$. For any pair $(i,j)\in\mathcal{E}$ and any input $x$, every first‑order stationary point of $\mathcal{L}$ satisfies
\begin{align}
    s_i(x)\;\le\;0 \quad\text{or}\quad s_j(x)\;\le\;0
\end{align} 
\end{proposition}
The proof is deferred to~\Cref{app:stationary_proof}. Intuitively,~\Cref{prop:exclusive_stationary} establishes that the regularization prevents any stationary solution where both logits in an exclusive pair are simultaneously positive, thereby enforcing the biological exclusive constraint. Furthermore, to demonstrate the necessity and advantages of our exclusive regularization, we show that the regularizer yields both a tighter generalization error bound and an explicit guarantee on conflict probability.

\begin{theorem}[Generalization and Exclusivity Guarantee]
\label{thm:gen-gap-conflict}
Assume bounded inputs and linear logits as in Assumptions \ref{asm:A1} and \ref{asm:A2} (see ~\Cref{app:assumptions}).  
Fix $\lambda>0$ and let
$\hat R_{\!S}(W)$ be the empirical risk on a sample
$S=\{(\mathbf{x}_n,\mathbf{y}_n)\}_{n=1}^N$,
drawn i.i.d.\ from $\mathcal D$, and let $W^*_\lambda:=\arg\min_W \hat R_{\!S}(W)$.  
Define $C_\lambda:=\hat R_{\!S}(W^*_\lambda)/\lambda$.  
Then, with probability at least $1-\delta$ over the draw of the training sample $\mathcal{S}$, we have the following \textbf{(a) generalization gap} and \textbf{(b) conflict probability}:
\begin{align}
&\textbf{(a):}\:
R(W)
\leq
\widehat{R}_{\mathcal{S}}(W)
+
\frac{2C_\lambda}{\sqrt{N}}
+
3\sqrt{\frac{\log(2/\delta)}{2N}}
\\
&\textbf{(b):}
\!\!\!\Pr_{(\mathbf{x},\mathbf{y})\sim\mathcal{D}}\!\bigl(s_i(\mathbf{x})\!>\!0,s_j(\mathbf{x})\!>\!0\bigr)\!\leq\! \frac{R(W)}{\lambda\,\log 2},
\forall\,(i,j)\!\in\!\mathcal{E}
\end{align}
\end{theorem}
The detailed proof can be found in~\Cref{app:generalization_proof}. Part (a) of~\Cref{thm:gen-gap-conflict} indicates the exclusive regularization $\mathcal{L}_{\mathrm{exc}}$ functions as an implicit norm regularizer, ensuring a generalization gap that is at least as tight as that of the unregularized loss $\mathcal{L}_{\mathrm{MLC}}$. Part (b) further shows that the probability of predicting conflicting labels decays at least as fast as the generalization gap—a form of control not offered by plain loss $\mathcal{L}_{\mathrm{MLC}}$. Together, these results provide a rigorous guarantee that our proposed objective can be broadly applied to ontology classification tasks with mutually exclusive label pairs.

\begin{table*}[!t]
\centering

\small
\begin{tabular}{l|l|cc|cc|cc|cc|cc}
\toprule
\multicolumn{2}{c|}{\textbf{Phenotype Frequency}} &
\multicolumn{2}{c|}{\textbf{11–30}} & 
\multicolumn{2}{c|}{\textbf{31–100}} & 
\multicolumn{2}{c|}{\textbf{101–300}} & 
\multicolumn{2}{c|}{\textbf{$\geq$301}} &
\multicolumn{2}{c}{\textbf{All}} \\
\cmidrule(lr){1-2} \cmidrule(lr){3-4} \cmidrule(lr){5-6} \cmidrule(lr){7-8} \cmidrule(lr){9-10} \cmidrule(lr){11-12}
\textbf{Dataset} & \textbf{Method} 
& \textbf{$F_{max}$~$\uparrow$} & AUC~$\uparrow$ 
& \textbf{$F_{max}$~$\uparrow$} & AUC~$\uparrow$
& \textbf{$F_{max}$~$\uparrow$} & AUC~$\uparrow$
& \textbf{$F_{max}$~$\uparrow$} & AUC~$\uparrow$ 
& \textbf{$F_{max}$~$\uparrow$} & AUC~$\uparrow$ \\
\midrule

MPO & max-BLAST & 8.14 & 50.61 & 10.96 & 50.52 & 13.92 & 50.55 & 21.78 & 50.79 & 16.65 & 50.59 \\
& w-BLAST & 11.78 & 52.74 & 13.59 & 53.16 & 17.23 & 53.71 & 28.02 & 53.46 & 22.73 & 53.10 \\
& kmer2Vec+LR & 9.82 & 55.63 & 11.46 & 55.36 & 16.78 & 55.57 & 33.74 & 54.91 & 28.75 & 55.45 \\
& $*$DeepGoPlus & 1.42 & 52.58 & 8.56 & 52.52 & 11.79 & 52.74 & 33.26 & 52.53 & 28.51 & 52.58 \\
& $*$SPROF-GO  & 5.43 & 52.73 & 6.59 & 52.13 & 11.22 & 51.71 & 20.42 & 50.92 & 12.33 & 52.06 \\
& $*$InterLabelGO & \underline{11.90} & \underline{59.27} & 13.19 & 57.36 & 17.03 & 56.03 & \underline{34.15} & 54.29 & \underline{29.66} & 57.61 \\
& DeepPheno & 8.45 & 58.09 & \underline{13.98} & \underline{64.05} & \underline{18.03} & \underline{61.94} & 33.84 & \underline{60.59} & 28.80 & \underline{59.89}\\
\cmidrule(lr){2-12}
& \textbf{GenePheno} & \textbf{13.86} & \textbf{68.02} & \textbf{16.12} & \textbf{69.73} & \textbf{19.49} & \textbf{68.09} & \textbf{36.06} & \textbf{61.43} & \textbf{31.14} & \textbf{67.86} \\
\midrule
\midrule
HPO & max-BLAST & 14.71 & 51.25 & 17.72 & 51.83 & 21.51 & 52.27 & 35.74 & 52.43 & 25.77 & 51.74 \\
& w-BLAST & 16.19 & 53.65 & 19.76 & 54.40 & 24.12 & 54.81 & 40.67 & 54.82 & 31.03 & 54.22 \\
& kmer2Vec+LR & 15.08 & 54.02 & \underline{20.83} & 54.97 & 26.32 & 54.62 & 48.58 & 54.52 & 40.11 & 54.47 \\
& $*$DeepGoPlus & 5.86 & 51.33 & 16.94 & 51.02 & 25.25 & 51.03 & 48.71 & 51.88 & 40.18 & 51.24 \\
& $*$SPROF-GO & 11.96 & 51.98 & 15.75 & 51.67 & 23.27 & 50.65 & 35.42 & 52.11 & 23.71 & 51.63 \\
& $*$InterLabelGO & 16.04 & \underline{57.52} & 19.13 & \underline{58.40} & 26.09 & 56.96 & \underline{49.17} & 58.39 & \underline{40.45} & 57.84 \\
& DeepPheno & \underline{17.48} & 55.68 & 18.73 & 58.32 & \underline{27.08} & \underline{60.59} & 49.02 & \underline{63.03} & 40.37 & \underline{58.17}\\
\cmidrule(lr){2-12}
& \textbf{GenePheno} & \textbf{22.02} & \textbf{72.79} & \textbf{26.34} & \textbf{73.09} & \textbf{31.65} & \textbf{69.03} & \textbf{51.95} & \textbf{65.82} & \textbf{43.17} & \textbf{71.44} \\
\midrule
\midrule
GWAS & max-BLAST & 1.31 & 49.84 & 6.93 & 50.32 & 13.69 & 49.98 & 28.49 & 48.59 & 23.15 & 49.95 \\
& w-BLAST & 5.42 & 50.35 & 14.95 & 52.47 & 23.00 & 51.58 & 39.36 & 53.10 &31.08& 51.54 \\
& kmer2Vec+LR & 9.07 & \textbf{57.64} & 14.35 & 50.16 & 25.64 & 51.35 & 46.75 & \underline{54.36} & 35.68 & \underline{53.53} \\
& $*$DeepGoPlus & 7.16 & 52.95 & 13.12 & \underline{54.97} & 26.37 & 52.16 & \underline{47.83} & 51.80 & 37.24 & 53.45 \\
& $*$SPROF-GO & 6.70 & 51.01 & 9.40 & 52.24 & 26.69 & 53.36 & 38.35 & 53.01 & 31.08 & 52.03 \\
& $*$InterLabelGO & \underline{9.38} & 50.90 & \underline{16.05} & 53.91 & 25.62 & \underline{53.63} & 44.84 & 49.80 & \underline{37.83} & 52.37 \\
& DeepPheno & 9.07 & 50.14 & 14.12 & 53.47 & \underline{26.97} & 52.96 & 46.92 & 53.74 & 36.87 & 52.12\\
\cmidrule(lr){2-12}
& \textbf{GenePheno} & \textbf{9.50} & \underline{56.64} & \textbf{17.03} & \textbf{55.44} & \textbf{27.22} & \textbf{56.47} & \textbf{48.57} & \textbf{57.75} & \textbf{40.54} & 
\textbf{56.34} \\
\midrule
\midrule
CAFA2& max-BLAST & 14.35 & 50.16 & 16.97 & 49.93 & 20.92 & 49.65 & 31.39 & 49.77 & 20.21 & 49.94 \\
wPPI& w-BLAST & 16.50 & 51.37 & 21.10 & 50.98 & 26.73 & 51.30 & 39.23 & 51.79 & 26.37 & 51.24 \\
& kmer2Vec+LR & 17.28 & 54.61 & 24.29 & 52.98 & 30.14 & 51.58 & 45.41 & 52.02 & 36.07 & 53.13 \\
& $*$DeepGoPlus & 12.95 & 51.84 & 16.58 & 54.59 & 28.01 & 53.49 & 46.69 & 52.59 & 37.18 & 53.32 \\
& $*$SPROF-GO & 15.33 & 50.04 & 20.69 & 49.78 & 22.43 & 47.17 & 41.08 & 47.11 & 21.30 & 49.24 \\
& $*$InterLabelGO & 16.07 & 54.37 & 19.55 & 51.75 & 26.69 & 52.73 & 45.02 & 55.46 & 35.47 & 53.31 \\
& DeepPheno & 15.20 & 54.62 & 20.47 & 54.63 & \underline{30.93} & 57.44 & \underline{47.37} & 56.37 & \underline{37.55} & 55.35 \\
& GraphPheno & 18.01 & \underline{56.35} & 24.68 & 59.04 & 31.88 & \underline{59.47} & 46.60 & \underline{57.13} & 36.02 & \underline{58.11} \\
& HPOFiller & 14.37 & 49.57 & 22.58 & 48.44 & 28.84 & 49.14 & 43.18 & 49.10 & 23.32 & 48.87 \\
& HPODNets & 18.35 & 49.88 & 22.46 & 49.28 & 26.07 & 49.16 & 43.17 & 48.95 & 31.21 & 49.39 \\
& SSLpheno & \underline{21.49} & 55.39 & \underline{26.57} & \underline{57.17} & 30.86 & 59.21 & 42.05 & 49.42 & 31.66 & 56.19 \\
\cmidrule(lr){2-12}
& \textbf{GenePheno} &\textbf{21.89} & \textbf{63.50} & \textbf{27.48} & \textbf{62.93} & \textbf{33.97} & \textbf{61.27} & \textbf{48.13} & \textbf{60.25} & \textbf{37.94} & \textbf{62.53} \\
\bottomrule
\end{tabular}
\caption{Phenotype prediction results on four datasets: MPO, HPO, GWAS, and CAFA2 (wPPI).
Baselines marked with $*$ are adapted from protein sequence–to–GO function prediction models to gene sequence–to–phenotype prediction. Results are stratified by phenotype label frequency into four intervals: 11–30, 31–100, 101–300, and $\geq$301. “All” indicates performance on the full dataset. Percentage scores are reported for $F_{\max}$ and AUC. Bolded values denote the highest score in each column; underlined values indicate the second-best.}
\label{tab:dna-to-phenotype-stacked}
\end{table*}

\subsection{Mechanism-aware Learning with GO}
Our regularized contrastive loss establishes the relationship between genetic information encoded in gene sequences and observable phenotypes. To further capture the biological mechanisms underlying phenotype formation, we incorporate GO functions as an additional modality through a dual-granularity approach. 

Formally, for each gene $g_i$, we extract two complementary GO representations. First, we obtain fine-grained GO annotations with clear experimental evidence from UniEntrezDB~\citep{miao2024unientrezdb} and process them through GoBERT to generate embeddings $\mathbf{h}_i$ that capture implicit functional correlations. These embeddings undergo cross-attention with gene sequence information to facilitate modality fusion. Second, we derive coarse-grained GO functions by propagating the fine-grained annotations along ``is\_a'' and ``part\_of'' relationships~\citep{gaudet2017primer} in the GO DAG, resulting in a binary vector $\mathbf{g}_i\in\{0,1\}^{n}$ representing general biological mechanisms.

We designate the second-last layer of our network as a GO bottleneck, where outputs from $n$ specific nodes serve as predictions $\widehat{\mathbf{g}}_i$ for the coarse-grained GO annotations. This bottleneck is optimized using another contrastive loss $\mathcal{L}_\mathrm{MLC}^{\mathrm{GO}}(\widehat{\mathbf{g}}, \mathbf{g})$. During inference, the learned connection weights $w_{ij}$ between GO node $\mathbf{g}_i$ and phenotype $\mathbf{y}_j$ quantify function-phenotype associations. The dual role of this bottleneck layer is significant: it not only introduces biologically-informed guidance throughout the end-to-end learning process, but also provides interpretable insights into the general biological mechanisms underlying phenotype formation during inference.

\subsection{Overall Learning Framework}
The architecture of our proposed model is illustrated in~\Cref{fig:our_model}. Our framework takes gene sequence embeddings $\mathbf{e}_i$ and GO embeddings $\mathbf{h}_i$ as input, facilitating information exchange between these modalities through cross-attention mechanism. These representations are subsequently integrated into a multi-modal embedding $\mathbf{x}_i$ via attention-pooling and concatenation. The final phenotype predictions are formulated as $\widehat{\mathbf{y}}_i = f_\psi(\mathbf{x_i}) = f_\psi(f_\theta(\mathbf{e}_i, \mathbf{h}_i))$ where $f_\theta(\cdot, \cdot)$ implements the cross-attention mechanism and $f_\psi(\cdot)$ includes MLP and GO bottleneck layer. Additionally, we derive exclusive label pairs set $\mathcal{E}$ using LLM and designate specific activations from intermediate layers as GO prediction outputs $\widehat{\mathbf{g}}_i$ along with corresponding ground-truth $\mathbf{g}_i$. Integrating these components, we formulate our comprehensive minimization objective as:
\begin{align}
\mathcal{L} = \mathcal{L}_\mathrm{MLC}(\mathbf{\widehat{y}, y}) + \lambda_1 \mathcal{L}_\mathrm{ex}(\widehat{\mathbf{y}}, \mathbf{y}, \mathcal{E}) + \lambda_2 \mathcal{L}_\mathrm{MLC}^{\mathrm{GO}}(\widehat{\mathbf{g}}, \mathbf{g})
\end{align}
where $\lambda_1$ and $\lambda_2$ are hyperparameters controlling the contribution of exclusivity and bottleneck loss terms, respectively.

\section{Experiments}
\label{sec:experiments}
We perform a comprehensive evaluation of GenePheno on four curated gebe sequence–to–phenotype datasets: HPO, MPO, GWAS, and CAFA2 wPPI. GenePheno is implemented with GENERator~\citep{wu2025generator} as the gene sequence encoder and GoBERT~\citep{miao2025gobert} as the GO function encoder. 

We compare GenePheno against several baselines, including max and weighted BLAST-based methods~\citep{camacho2009blast+}, kmer2vec~\citep{ren2022kmer2vec} combined with logistic regression, and three adapted sequence-based gene function prediction models: DeepGOPlus~\citep{kulmanov2020deepgoplus}, SPROF-GO~\citep{yuan2023fast}, and InterLabelGO~\citep{liu2024interlabelgo+}.

As many genes lack experimentally validated PPI and GO annotations, we construct the CAFA2 wPPI dataset by intersecting the widely used CAFA2 challenge set~\citep{jiang2016expanded} with established PPI networks~\citep{szklarczyk2015string, franz2018genemania, hwang2019humannet}. All genes in the resulting dataset have valid PPI information and GO annotations, enabling fair comparison with methods that rely on curated genetic modalities. On this dataset, we further evaluate curated-modality-based phenotype prediction methods, including DeepPheno~\citep{kulmanov2020deeppheno}, GraphPheno~\citep{liu2022integration}, HPOFiller~\citep{liu2021hpofiller}, HPODNets~\citep{liu2022hpodnets}, and SSLPheno~\citep{bi2023sslpheno}.

We follow CAFA-style preprocessing for all datasets and adopt evaluation protocols from prior work~\cite{cho2016compact, liu2022hpodnets, bi2023sslpheno}, stratifying results by phenotype label frequency. Performance is reported using two widely accepted metrics: gene-centric $F_{\max}$ and phenotype-centric AUC~\cite{jiang2016expanded}. More details on dataset construction, baseline adaptation, experimental setup and reproducibility details are provided in the~\Cref{app:dataset_collection,app:evaluation_metrics,app:baseline_adaptation,app:experiment_details,app:reproducibility}.

\section{Results}
We report results across phenotype frequency ranges using gene‐centric $F_{\max}$ and phenotype‐centric AUC to compare GenePheno against baseline methods. We perform an ablation study to quantify each module’s contribution and present case studies showing that the GO bottleneck layer captures generalizable mechanisms of phenotype formation.

\begin{figure*}[!t]
    \centering
\includegraphics[width=1\textwidth]{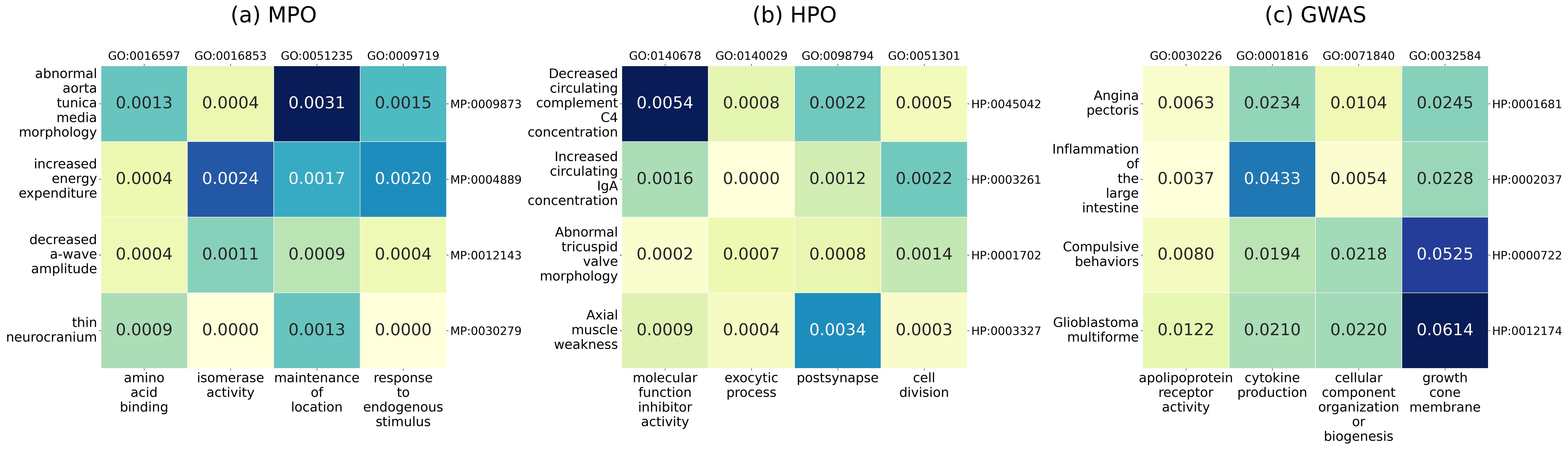}  
    \caption{Sample bottleneck weight heatmap of MPO, HPO, and GWAS datasets. Darker colors indicate that the phenotype and GO function have a higher correlation.} 
    \label{fig:case_study}
\end{figure*}

\begin{table*}[t]
\centering
\small
\begin{tabular}{l|c|cc|cc|cc|cc}
\toprule

& \textbf{Interpretable} & \multicolumn{2}{c|}{\textbf{HPO}} 
& \multicolumn{2}{c|}{\textbf{MPO}} 
& \multicolumn{2}{c|}{\textbf{GWAS}}
& \multicolumn{2}{c}{\textbf{CAFA2 (wPPI)}} \\
\cmidrule(lr){3-4} \cmidrule(lr){5-6} \cmidrule(lr){7-8} \cmidrule(lr){9-10}
\textbf{Method} & \textbf{Mechanism} & \textbf{$F_{max}$~$\uparrow$} & AUC~$\uparrow$ 
& \textbf{$F_{max}$~$\uparrow$} & AUC~$\uparrow$
& \textbf{$F_{max}$~$\uparrow$} & AUC~$\uparrow$
& \textbf{$F_{max}$~$\uparrow$} & AUC~$\uparrow$ \\
\midrule
w/o GO Input & \ding{51} & 39.96 & 56.60 & 28.21 & 54.19 & 36.82 & 52.59 & 35.65 & 53.83 \\
w/o Sequence Input & \ding{51} & 40.99 & 60.23 & 29.13 & 64.72 & 36.23 & 51.10 & 36.39 & 59.33 \\
w/o Contrastive Loss & \ding{51} & 41.83 & 65.75 & 29.99 & 64.64 & 36.05 & 52.25 & 37.06 & 50.86 \\
w/o Mutual Exclusive Loss & \ding{51} & 43.07 & 65.22 & 29.92 & 62.92 & 39.04 & 56.01 & 37.13 & 61.04 \\
w/o Bottleneck Loss & \ding{55} & 42.94 & 68.17 & 30.13 & 64.67 & 39.02 & 54.36 & 36.59 & 61.39 \\
\midrule
Complete Model & \ding{51} & \textbf{43.17} & \textbf{71.44} & \textbf{31.14} & \textbf{67.86} & \textbf{40.54} & \textbf{56.34} & \textbf{37.94} & \textbf{62.53} \\
\bottomrule

\end{tabular}
\caption{Ablation study results. Results are reported under the ``All'' setting for four datasets.}
\label{tab:ablation-studies}
\end{table*}

\subsection{Benchmarking results}

\Cref{tab:dna-to-phenotype-stacked} presents comprehensive experimental results across four phenotype-prediction datasets. GenePheno consistently outperforms all baseline methods, achieving substantial improvements in both $F_{\max}$ and AUC metrics. On the larger-scale datasets, GenePheno demonstrates particularly strong performance: for MPO, it surpasses the second-best method by +1.48 in $F_{\max}$ and +7.97 in AUC, while on HPO, the margins increase to +2.72 in $F_{\max}$ and +13.27 in AUC. These significant improvements can be attributed to the contrastive learning objective, which effectively captures inter-label relationships in high-dimensional phenotype spaces.

GenePheno maintains robust performance even on smaller datasets with limited genes and phenotypes. On GWAS, it improves performance from 37.83/53.53 ($F_{\max}$/AUC) achieved by the best baseline to 40.54/56.34, while consistently maintaining a 0.8–2.4-point advantage in high-frequency ($\geq$301) subsets. Similarly, on CAFA2, GenePheno demonstrates strong generalization capabilities under sparse supervision, particularly proved by its performance on low-frequency bins. This consistent performance across diverse data regimes validates the effectiveness of integrating DNA and GO encoders with cross-phenotype attention mechanisms for scalable phenotype prediction. The label-correlation aware contrastive loss further enhances the model's ability to generalize from limited supervision. Overall, GenePheno achieves state-of-the-art performance across all evaluation settings, establishing itself as a unified and effective approach for phenotype prediction tasks.

\subsection{Ablation Studies}
We conduct ablation studies to assess the impact of key components in GenePheno (Table~\ref{tab:ablation-studies}). Removing the contrastive loss causes the largest drop in performance, highlighting its role in modeling phenotype correlations. GO inputs are also critical, as their removal notably reduces AUC. The mutual exclusivity and bottleneck losses consistently improve results, supporting model interpretability. Sequence input further adds complementary value, as its exclusion leads to a performance decline.

\subsection{Case studies}
We analyze GenePheno’s bottleneck weights to interpret phenotype formation. Representative links are shown in Figure~\ref{fig:case_study}, with detailed analysis in~\Cref{app:case_study}.
\paragraph{MPO}
GenePheno identifies biologically plausible associations in this mouse gene knockout dataset. Two representative cases are shown in~\Cref{fig:case_study}(a). First, ``maintenance of location'' (GO:0051235) is linked to ``abnormal aorta tunica media morphology'' (MP:0009873), a feature of aortic aneurysms caused by disorganized cell and matrix structure~\citep{tang2005hyperplastic}. Second, ``isomerase activity'' (GO:0016853) is associated with ``increased energy expenditure'' (MP:0004889), aligning with findings that prolyl isomerases like Pin1 modulate skeletal muscle metabolism via SERCA activity~\citep{nakatsu2024prolyl}.

\paragraph{HPO}
The HPO dataset curates gene–phenotype associations from medical literature~\citep{kohler2021human}. ~\Cref{fig:case_study}(b) shows sample bottleneck weights from this dataset. GenePheno reveals a strong association between ``molecular function inhibitor'' (GO:0140678) and ``decreased circulating complement C4 concentration'' (HP:0045042), consistent with studies showing that complement inhibitors reduce C4 levels~\citep{coss2023complement, garred2021therapeutic}. Another plausible link is found between ``postsynapse'' (GO:0098794) and ``axial muscle weakness'' (HP:0003327), as impaired signal transmission at the neuromuscular junction can lead to muscle weakness~\citep{jones2017cellular, hirsch2007neuromuscular}.

\paragraph{GWAS}
The GWAS dataset links genes to phenotypes through SNP associations~\citep{buniello2019nhgri, macarthur2017new}, and focuses on a limited set of disease-related traits compared to the broader coverage of HPO and MPO. GenePheno interpretation weights are shown in~\Cref{fig:case_study}(c). One highlighted function, ``growth cone membrane'' (GO:0032580), is involved in axonal navigation~\citep{vitriol2012growth, igarashi2019molecular}, and is strongly associated with ``compulsive behaviors'' (HP:0000722) and ``glioblastoma multiforme'' (HP:0012174). GenePheno also links ``cytokine production'' (GO:0001816) to ``inflammation of the large intestine'' (HP:0002037), consistent with the role of cytokines in coordinating immune responses~\citep{zhang2007cytokines, sanchez2008role, escalante2025leaky}.

\section{Conclusion}
We present GenePheno, the first end-to-end deep learning framework for directly mapping gene sequences to phenotypic traits. By integrating gene function information at both fine and coarse levels and modeling inter-phenotype dependencies and exclusivity, GenePheno bridges the modality gap between sequences and observable traits while providing interpretable, biologically grounded insights. To support this task, we curate four benchmark datasets and demonstrate strong, consistent performance through comprehensive experiments and case studies.

\section{Acknowledgements}
This work was partially supported by US National Science Foundation IIS-2412195, CCF-2400785, the Cancer Prevention and Research Institute of Texas (CPRIT) award (RP230363), the National Institutes of Health (NIH) R01 award (1R01AI190103-01) and Microsoft Accelerate Foundation Models Research (2024).

\bibliography{aaai2026}

Check whether the conference requires a reproducibility checklist to be included in the paper.
If so, you can uncomment the following line and adjust the path to include it.
\clearpage
\newpage
\appendix
\appendix
\setcounter{secnumdepth}{2}
\setcounter{page}{1}
\pagestyle{plain}
\setcounter{section}{0}         
\setcounter{equation}{0}
\renewcommand\thesection{\Alph{section}}

{\centering
\Large
Appendix \\}

\section{Assumptions for~\Cref{thm:gen-gap-conflict}}
\label{app:assumptions}

\begin{assumption}[Bounded input space]
\label{asm:A1}
The feature domain is a closed $\ell_2$–ball,   
\[
\mathcal{X}\;\subseteq\;\bigl\{\mathbf{x}\in\mathbb{R}^d : \|\mathbf{x}\|_2 \le X_{\max}\bigr\},
\]
and the training examples  
$\{(\mathbf{x}_n,\mathbf{y}_n)\}_{n=1}^N$ are drawn i.i.d.\ from an unknown distribution  
$\mathcal{D}$ over $\mathcal{X}\times\{0,1\}^C$.
\end{assumption}

\begin{assumption}[Linear hypothesis class]
\label{asm:A2}
Each label $c\in\{1,\dots,C\}$ is scored by a linear logit  
\[
s_c(\mathbf{x}) \;=\; \mathbf{w}_c^{\top}\mathbf{x},
\quad
\mathbf{w}_c \in \mathbb{R}^d,
\]
and we write the parameter matrix as  
\(W=[\mathbf{w}_1,\dots,\mathbf{w}_C]\in\mathbb{R}^{d\times C}\).
No explicit norm constraint on $W$ is required beyond what appears in the
generalization bound.
\end{assumption}

\section{Proof of~\Cref{eq:multi-label_loss}}
\label{app:loss_derivation}

Recall that the original InfoNCE loss is defined as
\begin{align}
\label{eq:original_infonce}
\mathcal{L}_{\text{InfoNCE}}
  = -\log
    \frac{\exp\bigl(z_i \cdot z_j / \tau\bigr)}
         {\sum_{k=1}^{N}\mathds{1}_{[k\neq i]}\,
          \exp\bigl(z_i \cdot z_k / \tau\bigr)}
\end{align}
where \(z_i\) denotes the embedding of the \(i\)-th input.  
This objective drives the similarity (inner product) of the positive pair \((z_i,z_j)\) to exceed that of every positive–negative pair \((z_i,z_k)\).  
For multi-label classification, we extend the loss to accommodate multiple positive anchors.

The resulting multi-positive InfoNCE loss is
\begin{equation}
\mathcal{L}_{\mathrm{NCE}}
  = -\sum_{i\in\Omega_{\mathrm{+}}}
      \log
      \frac{\exp\bigl(s_i/\tau\bigr)}
           {\exp\bigl(s_i/\tau\bigr)
            +\sum_{j\in\Omega_{\mathrm{-}}}\exp\bigl(s_j^{(-)}/\tau\bigr)}
\end{equation}
We treat the prediction scores of true labels as “positives’’ \(s_i/\tau\) and the scores of false labels as “negatives’’ \(s_j^{(-)}/\tau\); one convenient choice is \(s_j^{(-)}=-s_j\), so a large \(s_j\) implies very low similarity.

Rewriting the per-positive term for a fixed \(i\) gives
\begin{align}
-\!\log\!\!\left(\!\!\frac{e^{s_i/\tau}}
         {\,e^{s_i/\tau}
          \!\!+\!\sum_{j}e^{s_j^{(-)}/\tau}}\!\!\right)
  \!\!=\log\bigl(e^{s_i/\tau}
                \!+\!\!\sum_{j}e^{s_j^{(-)}/\tau}\bigr) \!-\!\!\;\frac{s_i}{\tau}
\end{align}
Summing over \(i\in\Omega_{\mathrm{+}}\) yields
\begin{equation}
\mathcal{L}_{\mathrm{NCE}}^{+}
  =\sum_{i\in\Omega_{\mathrm{+}}}
     \log\Bigl(e^{s_i/\tau}
                 +\sum_{j\in\Omega_{\mathrm{-}}}
                   e^{s_j^{(-)}/\tau}\Bigr)
   \;-\;
   \frac{1}{\tau}\sum_{i\in\Omega_{\mathrm{+}}} s_i
\end{equation}

Optimizing \(\mathcal{L}_{\mathrm{NCE}}^{+}\) pushes each positive score above the negative scores.  
To symmetrically repel negatives from positives, we also swap roles—treat each negative \(j\) as an “anchor’’ and the positives as “negatives.’’  
This produces a second InfoNCE-like term:
\begin{align}
\mathcal{L}_{\mathrm{NCE}}^{-}
  &=-\sum_{j\in\Omega_{\mathrm{-}}}
      \log\frac{e^{s_j^{(-)}/\tau}}
               {\,e^{s_j^{(-)}/\tau}
                +\sum_{i\in\Omega_{\mathrm{+}}}e^{s_i/\tau}}\\
  &=\sum_{j\in\Omega_{\mathrm{-}}}
      \log\Bigl(e^{s_j^{(-)}/\tau}
                  +\sum_{i\in\Omega_{\mathrm{+}}}e^{s_i/\tau}\Bigr)
    \;+\;
    \frac{1}{\tau}\sum_{j\in\Omega_{\mathrm{-}}} s_j^{(-)}
\end{align}

Our final objective for multi-label classification is therefore
\[
\mathcal{L}_{\mathrm{MLC}}
  = \mathcal{L}_{\mathrm{NCE}}^{+}
  + \mathcal{L}_{\mathrm{NCE}}^{-}.
\]

\section{Equivalence Proof with ZLPR Loss}
\label{app:zlpr_connection}
From~\Cref{app:loss_derivation}, our multi-label classification loss can be written as
\begin{align}
    \mathcal{L}_{\mathrm{MLC}} = &\mathcal{L}_{\mathrm{NCE}}^+ + \mathcal{L}_{\mathrm{NCE}}^- \\
    = &\sum_{i\in\Omega_{\mathrm{+}}} \log\Bigl(1+e^{-s_i/\tau}\sum_{j\in\Omega_{\mathrm{-}}}e^{s_j^{(-)}/\tau}\Bigr) \nonumber\\ & +  
    \sum_{j\in\Omega_{\mathrm{-}}} \log\Bigl(1+e^{-s_j^{(-)}/\tau}\sum_{i\in\Omega_{\mathrm{+}}}e^{s_i/\tau}\Bigr) \\
     = &\sum_{i\in\Omega_{\mathrm{+}}} \log\Bigl(1+\sum_{j\in\Omega_{\mathrm{-}}}e^{(s_j^{(-)}-s_i)/\tau}\Bigr) \nonumber\\  & +
    \sum_{j\in\Omega_{\mathrm{-}}} \log\Bigl(1+\sum_{i\in\Omega_{\mathrm{+}}}e^{(s_i-s_j^{(-)})/\tau}\Bigr)
\end{align}
Observe that setting \(\tau=1\) and \(s_j^{(-)}=-s_j\) makes this expression identical to the TLPR loss introduced by~\citep{su2022zlpr}. If we further apply a zero threshold and combine the two log–sum–exp denominators, we derive an equivalent form of the ZLPR loss:
\begin{align}
&\sum_{i\in\Omega_{\mathrm{+}}}\!
\log\Bigl(1 \!+ \!\!\sum_{j\in\Omega_{\mathrm{-}\!}}e^{s_j^{(-)}-s_i}\Bigr)
\!+\!\!\!
\sum_{j\in\Omega_{\mathrm{-}}}\!
\log\Bigl(1 \!+\!\! \sum_{i\in\Omega_{\mathrm{+}\!}}e^{s_i - s_j^{(-)}}\Bigr)\\
 =& \log\Bigl(1 + \sum_{i\in\Omega_{\mathrm{+}}}e^{-s_i}\Bigr)
+
\log \Bigl(1 + \sum_{j\in\Omega_{\mathrm{neg}}}e^{\,s_j}\Bigr)\\
=& \mathcal{L}_{\mathrm{ZLPR}}
\end{align}

\section{Proof on \Cref{prop:exclusive_stationary}}
\label{app:stationary_proof}
Write $y_k\in\{0,1\}$ for the ground‑truth indicator of label~$k$
and denote the positive/negative index sets by
$\Omega_{+}=\{k:y_k=1\}$ and $\Omega_{-}=\{k:y_k=0\}$.
For a scalar $u$, define
\[
\mathrm{softplus}(u)=\log(1+e^{u}),
\quad
\sigma(u)=\frac{1}{1+e^{-u}}
\]

Without loss of generality, we omit the temperature as in~\Cref{app:zlpr_connection}, our learning objective has component‑wise gradient
\begin{align}
g^{\mathrm{Z}}_t
=&\frac{\partial\mathcal{L}_{\mathrm{MLC}}}{\partial s_t} \\
=& -y_t\frac{e^{-s_t}}{1+\sum_{i\in\Omega_{+}}e^{-s_i}} \nonumber\\
&+ (1-y_t)\frac{e^{s_t}}{1+\sum_{j\in\Omega_{-}}e^{s_j}}
\in(-1,1)
\label{eq:z_grad}
\end{align}
For every exclusive pair $(i,j)\in\mathcal{E}$,
\(
\partial_{s_i}\mathrm{softplus}(s_i+s_j)=\sigma(s_i+s_j),
\)
and analogously for~$s_j$.  Hence the full gradient can be expressed in
\begin{equation}
g_i
=
g^{\mathrm{Z}}_i
+
\lambda\!\!\!
\sum_{j:(i,j)\in\mathcal{E}}\!\!\!
\sigma(s_i+s_j),
\quad
g_j
=
g^{\mathrm{Z}}_j
+
\lambda\!\!\!
\sum_{i:(i,j)\in\mathcal{E}}\!\!\!
\sigma(s_i+s_j)
\label{eq:full_grad}
\end{equation}

Fix a pair $(i,j)\in\mathcal{E}$ and suppose, for contradiction, that at a
stationary point we have $s_i>0$ and $s_j>0$.

\begin{itemize}
    \item \smallskip\noindent
\textit{Case} $(y_i,y_j)=(1,0)$\,:  
From \eqref{eq:z_grad},
$g^{\mathrm{Z}}_i<0$ and $g^{\mathrm{Z}}_j>0$.
Since $s_i+s_j>0$, $\sigma(s_i+s_j)\in(0,1)$, and
\eqref{eq:full_grad} gives $g_j>0$.  Thus $g_j\neq0$, contradicting stationarity.

    \item \smallskip\noindent
\textit{Case} $(y_i,y_j)=(0,1)$\,:  Symmetric to the previous case; now $g_i>0$.

    \item \smallskip\noindent
\textit{Case} $(y_i,y_j)=(0,0)$\,:  
Equation~\eqref{eq:z_grad} yields $g^{\mathrm{Z}}_i>0$ and $g^{\mathrm{Z}}_j>0$,
so \eqref{eq:full_grad} implies $g_i,g_j>0$, again contradicting stationarity.
\end{itemize}

Because the label configuration $(y_i,y_j)=(1,1)$ is ruled out by definition of
$\mathcal{E}$, these three cases exhaust all possibilities.  In each one,
the assumption $s_i>0\ \wedge\ s_j>0$ violates the condition $g=0$.
Therefore a stationary point cannot have both logits positive, proving the proposition.

\section{Proof on \Cref{thm:gen-gap-conflict}}
\label{app:generalization_proof}

\subsection{Proof of Part~(a)}

\textbf{Weight-norm contraction.}
Fix any mutually exclusive pair \((i,j)\in\mathcal{E}\) and any example \((x,y)\).
If \(s_{i}(x)+s_{j}(x)>0\), then 
\[
\phi(s_{i},s_{j})
\coloneqq
\log\bigl(1+e^{s_{i}+s_{j}}\bigr)
\ge
s_{i}+s_{j}-\log 2
\]
Because $W^*_\lambda$ minimizes the empirical objective
\(\widehat{R}_{\mathcal{S}}\), subtracting a constant cannot change
the minimizer.  Hence
\begin{equation}
\label{eq:pair_constraint}
s_{i}(x^{(n)})+s_{j}(x^{(n)})
\le\frac{\widehat{R}_{\mathcal{S}}(W^*_\lambda)}{\lambda}
\coloneqq\;C_{\lambda},
\quad
\forall n\in[N]
\end{equation}

Since \(\|x^{(n)}\|_{2}\le X_{\max}\) and
\(s_{i}(x^{(n)})+s_{j}(x^{(n)})
      =(w_{i}+w_{j})^{\top}x^{(n)}\),
the Cauchy–Schwarz inequality yields
\(\|w_{i}+w_{j}\|_{2}\le C_{\lambda}/X_{\max}\).
Every label participates in at least one exclusive pair in \(\mathcal{E}\)
(or such a pair can always be constructed),
so repeated applications of the triangle inequality give
\begin{equation}
\label{eq:radius_bound}
\|w_{\ell}\|_{2}
\le\frac{C_{\lambda}}{X_{\max}},
\quad
\forall\ell\in\{1,\dots,L\}.
\end{equation}

Thus $W^*_\lambda$ lies in the Frobenius ball
\(\mathcal{B}_{2}\bigl(C_{\lambda}/X_{\max}\bigr)\),
whereas the unconstrained contrastive-MLC objective provides no such
control.  As \(\lambda\) increases, \(C_{\lambda}\) decreases
monotonically, shrinking the ball.

\medskip
\textbf{Lipschitz constant of the loss.}
From~\Cref{eq:radius_bound},
\(\max_{\ell}\|w_{\ell}\|_{2}\le C_{\lambda}/X_{\max}\),
so each score map \(x\mapsto s_{\ell}(x)\) is
\(C_{\lambda}\)-Lipschitz.
Both the contrastive-MLC term and \(\phi\) are 1-Lipschitz in their logits;
consequently the full loss \(\mathcal{L}\) is
\(C_{\lambda}\)-Lipschitz with respect to \(x\).

\medskip
\textbf{Rademacher complexity.}
Let \(\mathfrak{R}_{N}\) denote the empirical Rademacher complexity of the
function class
\[
\mathcal{F}_{\lambda}=
\bigl\{(x,y)\mapsto\mathcal{L}(W;(x,y))
  :W\in\mathcal{B}_{2}(C_{\lambda}/X_{\max})\bigr\}.
\]
Standard bounds for Lipschitz linear predictors
(Lemma 26.10 of~\citep{shalev2014understanding}) give
\(\mathfrak{R}_{N}\le C_{\lambda}/\sqrt{N}\).

\medskip
\paragraph{Convergence bound for the regularized loss.}
Define the population and empirical risks as
\begin{align}
R(W)   &\coloneqq\mathbb{E}_{(x,y)\sim\mathcal{D}}
          \bigl[\mathcal{L}\bigl(W;(x,y)\bigr)\bigr],\\[2pt]
R_S(W) &\coloneqq\frac{1}{N}\sum_{n=1}^{N}
          \mathcal{L}\bigl(W;(x^{(n)},y^{(n)})\bigr)
\end{align}
A standard Rademacher-complexity generalization theorem ensures that,
for any \(W\in\mathcal{F}_{\lambda}\) and with probability at least
\(1-\delta\) over the sample \(S\),
\begin{align}
    R(W)
\le
R_S(W)
+
2\,\mathfrak{R}_{N}\bigl(\mathcal{F}_{\lambda}\bigr)
+
3\sqrt{\frac{\log(2/\delta)}{2N}}
\end{align}
Substituting the bound
\(
\mathfrak{R}_{N}\bigl(\mathcal{F}_{\lambda}\bigr)
      \le C_{\lambda}/\sqrt{N}
\)
yields the explicit gap
\begin{align}
R(W)\le R_S(W)
+
\frac{2C_{\lambda}}{\sqrt{N}}
+
3\sqrt{\frac{\log(2/\delta)}{2N}}
\end{align}

Because \(C_{\lambda}\) decreases monotonically with the exclusivity
weight~\(\lambda\), the bound tightens as the penalty strengthens.

\begin{table*}[!t]
\centering

\begin{tabular}{llc}
\hline
\textbf{Phenotype 1} & \textbf{Phenotype 2} & \textbf{Conflict} \\
\hline
Abnormality of facial adipose tissue & Abnormality of subcutaneous fat tissue & 0 \\
Abnormality of facial adipose tissue & Lipodystrophy & 0 \\
Abnormality of facial adipose tissue & Increased adipose tissue & 0 \\
Abnormality of facial adipose tissue & Panniculitis & 0 \\
Abnormality of facial adipose tissue & Diet-resistant subcutaneous adipose tissue & 0 \\
Abnormality of facial adipose tissue & Adipocyte hypertrophy & 0 \\
Abnormality of facial adipose tissue & Lower extremity subcutanous fat hypertrophy & 0 \\
Abnormality of facial adipose tissue & Decreased adipose tissue & 0 \\
Abnormality of facial adipose tissue & Neoplasm of fatty tissue & 0 \\
Abnormality of subcutaneous fat tissue & Lipodystrophy & 0 \\
Abnormality of subcutaneous fat tissue & Increased adipose tissue & 0 \\
Abnormality of subcutaneous fat tissue & Panniculitis & 0 \\
Abnormality of subcutaneous fat tissue & Diet-resistant subcutaneous adipose tissue & 0 \\
Abnormality of subcutaneous fat tissue & Adipocyte hypertrophy & 0 \\
Abnormality of subcutaneous fat tissue & Lower extremity subcutanous fat hypertrophy & 0 \\
Abnormality of subcutaneous fat tissue & Decreased adipose tissue & 0 \\
Abnormality of subcutaneous fat tissue & Neoplasm of fatty tissue & 0 \\
Lipodystrophy & Increased adipose tissue & 1 \\
Lipodystrophy & Panniculitis & 0 \\
Lipodystrophy & Diet-resistant subcutaneous adipose tissue & 1 \\
Lipodystrophy & Adipocyte hypertrophy & 1 \\
Lipodystrophy & Lower extremity subcutanous fat hypertrophy & 1 \\
Lipodystrophy & Decreased adipose tissue & 0 \\
Lipodystrophy & Neoplasm of fatty tissue & 0 \\
Increased adipose tissue & Panniculitis & 0 \\
Increased adipose tissue & Diet-resistant subcutaneous adipose tissue & 0 \\
Increased adipose tissue & Adipocyte hypertrophy & 0 \\
Increased adipose tissue & Lower extremity subcutanous fat hypertrophy & 0 \\
Increased adipose tissue & Decreased adipose tissue & 1 \\
Increased adipose tissue & Neoplasm of fatty tissue & 0 \\
Panniculitis & Diet-resistant subcutaneous adipose tissue & 0 \\
Panniculitis & Adipocyte hypertrophy & 0 \\
Panniculitis & Lower extremity subcutanous fat hypertrophy & 0 \\
Panniculitis & Decreased adipose tissue & 0 \\
Panniculitis & Neoplasm of fatty tissue & 0 \\
Diet-resistant subcutaneous adipose tissue & Adipocyte hypertrophy & 0 \\
Diet-resistant subcutaneous adipose tissue & Lower extremity subcutanous fat hypertrophy & 0 \\
Diet-resistant subcutaneous adipose tissue & Decreased adipose tissue & 1 \\
Diet-resistant subcutaneous adipose tissue & Neoplasm of fatty tissue & 0 \\
Adipocyte hypertrophy & Lower extremity subcutanous fat hypertrophy & 0 \\
Adipocyte hypertrophy & Decreased adipose tissue & 1 \\
Adipocyte hypertrophy & Neoplasm of fatty tissue & 0 \\
Lower extremity subcutanous fat hypertrophy & Decreased adipose tissue & 1 \\
Lower extremity subcutanous fat hypertrophy & Neoplasm of fatty tissue & 0 \\
Decreased adipose tissue & Neoplasm of fatty tissue & 0 \\
\hline
\end{tabular}
\caption{Mutual exclusivity predictions for phenotype pairs under the group ``Abnormal adipose tissue morphology'' in HPO, 1 indicates the phenotypes are mutually exclusive, othervise 0.}
\label{app:exclusive_example}
\end{table*}

\begin{table}[!t]
\centering
\renewcommand{\arraystretch}{1.2}
\begin{tabular}{|p{0.95\linewidth}|}
\hline
You are a biomedical expert. Given a list of phenotype pairs, determine whether each pair is biologically mutually exclusive. \\
Return 1 if they are mutually exclusive (i.e., cannot occur together in the same DNA individual), or 0 if they are not exclusive or unclear. \\
\\
Respond in the format: \\
Phenotype\_1 vs Phenotype\_2: 1 or 0 \\
\hline
\end{tabular}
\caption{Prompt used for GPT-based mutual exclusivity annotation.}
\label{app:llm_prompt}
\end{table}

\subsection{Proof on Part (b)}

Because $u\mapsto\log(1+e^{u})$ is monotonically increasing and
$\log(1+e^{0})=\log 2$, the $\text{RHS}\ge1$ holds whenever both
$s_i,s_j>0$ and non-negative otherwise, hence the following holds for any logits $s_i,s_j\in\mathbb{R}$:
\begin{align}
\label{eq:conflict_bound}
    \mathds{1}(s_i>0 \wedge s_j>0)
      \le
      \frac{\log\bigl(1+e^{\,s_i+s_j}\bigr)}{\log 2}
\end{align}
Let $E_{ij}=\{s_i(x)>0\wedge s_j(x)>0\}$ and
$Z_{ij}(x)=\log\bigl(1+e^{s_i(x)+s_j(x)}\bigr)$.
By the inequality~\Cref{eq:conflict_bound},
\begin{align}
    \mathds{1}(E_{ij})
   \;\le\;
   \frac{Z_{ij}(x)}{\log 2}
\end{align}
Taking expectations over $\mathcal D$ gives
\begin{align}
    \Pr(E_{ij})
   \le
   \frac{\mathbb E_{\mathcal D}[Z_{ij}(x)]}{\log 2}
\end{align}
Because each penalty term appears in $\mathcal L$,
\begin{align}
    \lambda Z_{ij}(x)
   \le
   \mathcal L\bigl(W;(x,y)\bigr)
   \Longrightarrow
   \lambda\,\mathbb E_{\mathcal D}[Z_{ij}(x)]
   \le
   R(W)
\end{align}
Substituting back yields the stated inequality 
\begin{align}
    \Pr(E_{ij}) \le \frac{R(W)}{\lambda \log 2}
\end{align}

\section{Exclusive Phenotype Pairs Collection with LLM Prompt}
\label{app:llm_prompt}

Phenotype semantic exclusive pairs are only considered meaningful between first-order child nodes under the same phenotype ontology parent. Therefore, we first group all phenotype abnormalities from the ontology by their immediate parent term. Each group is named after its parent node (e.g., \texttt{Abnormal adipose tissue morphology}), and all possible unordered pairs between its children are enumerated as input to the LLM.

To identify mutually exclusive phenotype pairs, we use the Azure OpenAI API with a domain-specific prompt. For each group of phenotype pairs (stratified by ontology term clusters), we construct a text prompt describing the task and pass it to a deployed GPT model via the Azure OpenAI SDK. The model is instructed to label each pair with a binary flag: \texttt{1} if the phenotypes are biologically mutually exclusive (i.e., cannot occur in the same individual), and \texttt{0} otherwise.

Responses are parsed and saved per group. A 1-second delay is added between requests to avoid rate-limiting. All output files and errors are logged to the \texttt{OpenAI\_outputs\_hpo} directory. For each phenotype pair, the final output includes the original terms and the LLM-assigned exclusivity label.

In addition to LLM-based identification, we perform rule-based validation and supplementation using a hard-mapping strategy. If two phenotype terms are identical except for an opposing keyword (e.g., “increased” vs. “decreased”), they are also labeled as mutually exclusive. This provides a complementary check and adds robustness to the exclusivity identification process. The set of opposing keywords we consider includes:

\begin{itemize}
    \item increased vs. decreased
    \item high vs. low
    \item hyper vs. hypo
    \item macro vs. micro
    \item big vs. small
    \item tall vs. short
    \item over vs. under
    \item excess vs. deficient
    \item accelerated vs. delayed
    \item early vs. late
    \item enlarged vs. reduced
\end{itemize}

Both lowercase and capitalized variants of the keywords are included in matching logic.

The prompt format used in all API calls is shown in~\Cref{app:llm_prompt}. An example of the full annotation result for the group ``Abnormal adipose tissue morphology'' is provided in~\Cref{app:exclusive_example}.

In our datasets, phenotype annotations are ontology-based: the HPO dataset, GWAS dataset, and CAFA2 wPPI dataset use phenotype terms from the Human Phenotype Ontology (HPO), while the MPO dataset uses terms from the Mammalian Phenotype Ontology (MPO). Therefore, we identify mutually exclusive phenotype pairs separately for HPO and MPO.
For the HPO, we identify 5,830 such groups, generating 111,161 possible phenotype pairs, of which 6,687 were labeled as mutually exclusive. For the MPO, we analyze 5,795 groups with 63,400 phenotype pairs, resulting in 5,117 mutually exclusive pairs.

Incorporating mutually exclusive phenotype pairs in model training improves predictive performance. It enhances interpretability by enforcing biologically grounded constraints from phenotype ontologies, which discourages implausible phenotype combinations and reduces spurious correlations.

\begin{table*}[t]
\centering
\small
\renewcommand{\arraystretch}{1}

\begin{tabular}{lccccccc}
\hline
\textbf{Dataset} & \textbf{Phenotypes} & \textbf{Train/Valid} & \textbf{Test} & \textbf{Avg sequence Length} & \textbf{Label 1 Ratio} & \textbf{With GO} & \textbf{Avg GO Count} \\
\hline
CAFA2 wPPI & 6,149 & 2,312 & 269 & 2,018.31 & 0.0139 & 2,581 & 18.46 \\
HPO & 4,105 & 3,893 & 439 & 1,854.36 & 0.0333 & 4,275 & 17.36 \\
MPO & 4,700 & 11,880 & 1,048 & 10,907.90 & 0.0122 & 11,941 & 17.28 \\
GWAS & 155 & 4,814 & 410 & 1,942.29 & 0.0212 & 4,304 & 14.78 \\
\hline
\end{tabular}
\caption{Dataset statistics for CAFA2 wPPI, HPO, MPO, and GWAS. The ``Phenotypes" indicate the number of different phenotypes included in the multi-label classification. The ``Train/Valid” column indicates the number of samples used for model training and validation. “Test” denotes the number of genes in the test set. ``Label 1 Ratio” represents the proportion of positive phenotype labels across the entire dataset. ``With GO” refers to the number of samples with valid, manually reviewed GO annotations. ``Avg GO Count” is the average number of GO terms per gene among those with valid annotations. }
\label{app:dataset_stats}
\end{table*}

\section{Dataset Collection and Process}
\label{app:dataset_collection}
The datasets used in this study are curated from established biomedical sources. The HPO dataset is sourced from the official Human Phenotype Ontology database, which curates gene–phenotype associations from medical literature, covering diverse aspects of human phenotypes~\citep{kohler2021human}. The MPO dataset is derived from mouse gene knockout experiments, where phenotypic consequences are systematically observed following targeted gene disruptions~\citep{smith2009mammalian}. The GWAS dataset links genes to phenotypes through SNP associations curated from genome-wide association studies~\citep{buniello2019nhgri, macarthur2017new}. The CAFA2 wPPI dataset is constructed by intersecting the CAFA2 challenge gene set~\citep{jiang2016expanded} with widely used protein–protein interaction networks~\citep{szklarczyk2015string, franz2018genemania, hwang2019humannet}, ensuring that all included genes have valid PPI and GO annotations, allowing fair comparisons with methods reliant on curated modalities.
For the HPO, MPO, and GWAS datasets, we performed stratified splits based on phenotype label frequency to maintain balanced representation across both rare and common phenotypes. The split of the CAFA2 wPPI dataset follows the original CAFA2 challenge design. Gene sequences were obtained using reference genome annotations, and genes with coding sequences exceeding 60,000 base pairs were excluded due to model input length constraints.

To construct GO annotations, we used the UniEntrez database and retained only non-IEA (Inferred from Electronic Annotation) entries that are supported by experimental evidence and linked to publications. For each dataset, the bottleneck GO terms were selected by first propagating experimentally supported GO annotations along "is-a" and "part-of" relationships in the Gene Ontology hierarchy. We then selected GO terms at depth 2 (i.e., with shortest path length of 2 from the ontology root) as bottleneck targets. Any GO term that appeared as all-zero in the training set was removed from the bottleneck label set.

Following the CAFA evaluation protocol, we also propagated phenotype labels along ontology edges to reduce the impact of incomplete annotations and mitigate false negatives due to unobserved or unreported associations.

To allow fair comparison between methods that rely on different input modalities, we construct the CAFA2 wPPI dataset by intersecting the original CAFA2 challenge dataset~\citep{jiang2016expanded} with genes that are covered by the PPI networks used in baselines for curated modality-based prediction of phenotypes~\citep{liu2021hpofiller, liu2022hpodnets, bi2023sslpheno}. This ensures that all methods, including those that depend on PPI data, can be evaluated on a common subset of genes with complete input information. 

The statistics of all datasets used in this study are summarized in Table~\ref{app:dataset_stats}. Each dataset varies in size and structure: for example, GWAS contains the largest number of gene samples (12,928), while CAFA2 wPPI includes 2,581 genes. The average DNA sequence length ranges from 1,854 bp (HPO) to over 10,900 bp (MPO). We report the number of training and validation samples jointly, alongside a separate count for the test set to reflect the evaluation split. The ratio of positive phenotype labels is generally low (e.g., 1.2\% in GWAS), reflecting the sparsity typical of phenotype annotation tasks. Most datasets, except HPO, include samples with manually curated GO annotations. For genes with valid GO data, the number of associated GO terms averages between 14.8 (GWAS) and 18.5 (CAFA2 wPPI).

\section{Evaluation Metrics}
\label{app:evaluation_metrics} 
\subsection{Gene-centric $F_{max}$}
\label{app:evaluation_metrics_fmax} 
We evaluate prediction performance using the gene-centric $F_{\max}$ metric, following the protocol used in the CAFA2 challenge~\citep{jiang2016expanded}. For each gene, we first expand its ground-truth phenotype annotations by propagating them upward along the ontology hierarchy using “is-a” and “part-of” relationships. This step ensures that parent phenotypes are included, reflecting the structure of phenotype ontologies such as HPO and MPO.

The model produces a ranked list of phenotype predictions with prediction scores between 0 and 1. We then compute precision and recall across 100 thresholds (from 0 to 1 with a step size of 0.01), treating any phenotype with a predicted score above a threshold as positive. At each threshold, we calculate the average precision and recall across all genes, and compute their harmonic mean. The $F_{\max}$ is defined as the highest such F1 score over all thresholds. This metric accounts for the multi-label nature of phenotype prediction and evaluates how well the model captures the complete phenotype profile for each gene.

\subsection{Phenotype-centric $AUC$}
We also evaluate model performance using phenotype-centric AUC, which follows the term-centric evaluation protocol in CAFA2~\citep{jiang2016expanded}. For each phenotype term, we treat the prediction task as a binary classification problem across all genes and compute the area under the precision–recall curve (AUC). The final phenotype-centric AUC is reported as the average AUC across all phenotype terms. This metric reflects the model’s ability to correctly identify genes associated with each phenotype and is particularly informative for evaluating performance on rare or imbalanced phenotypes. Compared to gene-centric evaluation, phenotype-centric AUC highlights how well a model generalizes across diverse phenotype categories.

\section{Baseline Selection and Description}
\label{app:baseline_adaptation}

\subsection{max-BLAST}

We use sequence identity obtained from BLASTN~\citep{camacho2009blast+} as the similarity metric to construct a sequence similarity-based prediction method. For each test gene, we identify similar sequences in the training set using BLASTN, and transfer annotations from the most similar sequences. The prediction score for a test gene and a phenotype term is computed based on the normalized identity score across similar training sequences.

For each test gene $q$ and phenotype term $f$, the prediction score is calculated as:

\[
S(q, f) = \frac{\max_{s \in E} \text{identity}(q, s) \cdot \mathds{1}(f \in T_s)}{\max_{s \in E} \text{identity}(q, s)}
\]

where $E$ is the set of training sequences that are similar to $q$, filtered by e-value threshold. $T_s$ is the set of HPO terms annotated to training protein $s$, and $\mathds{1}(f \in T_s)$ is an indicator function that returns 1 if $s$ is annotated with $f$, and 0 otherwise.

This formulation assigns to each HPO term a score proportional to its presence among the most similar training sequences, normalized by the highest similarity score.
\subsection{w-BLAST}
We implement a weighted-BLAST baseline by aggregating phenotype annotations from all training sequences that exhibit high similarity to the test gene. Specifically, we perform BLASTN~\citep{camacho2009blast+} between each test DNA sequence and the training sequences, retaining only those hits with alignment length greater than 50 base pairs. For each matched training sequence, we collect its phenotype annotations and weight their contributions by the corresponding BLAST identity score.

Formally, for a test gene $q$ and a phenotype term $f$, the prediction score is computed as:

\[
S(q, f) = \frac{\sum\limits_{s \in E} \text{identity}(q, s) \cdot \mathds{1}(f \in T_s)}{\sum\limits_{s \in E} \text{identity}(q, s)}
\]

where $E$ is the set of training sequences aligned to $q$ with a matched length greater than 50 base pairs. $T_s$ is the set of phenotype terms annotated to sequence $s$, and $\mathds{1}(f \in T_s)$ is an indicator function that returns 1 if $f \in T_s$, and 0 otherwise.

This method assigns phenotype scores to each test gene based on weighted support from similar sequences, reflecting both annotation presence and sequence similarity.

\subsection{Kmer2Vec + LR}
Kmer2vec~\citep{ren2022kmer2vec} is a contextual alignment-free method for DNA sequence comparison that semantically embeds k-mers into word2vec vectors, capturing structural information within genomic sequences. We leverage Kmer2vec's pre-calculated embeddings, which represent DNA k-mers in a 300-dimensional vector space for k-mer lengths ranging from 3 to 8. For our multi-label phenotype prediction task, we specifically utilize the 3-mer embeddings to represent our input DNA sequences. These embeddings serve as feature vectors for a logistic regression model with L2 regularization (Ridge regression), which we employ to predict multiple phenotype labels simultaneously from the embedded DNA data.

\subsection{$*$DeepGoPlus}
DeepGOPlus~\citep{kulmanov2020deepgoplus} is originally designed to predict GO terms from protein sequences using convolutional neural networks. We adapt this model for DNA sequence–to–phenotype prediction by replacing the original amino acid one-hot encoding with a 3-mer one-hot encoding of the DNA sequence. The resulting 3-mer encoded DNA sequences are processed using the same convolutional architecture to produce phenotype predictions. The model is trained using binary cross-entropy over phenotype labels.

\subsection{$*$SPROF-GO}
SPROF-GO~\citep{yuan2023fast} predicts GO functions from protein sequences using ProtT5~\citep{elnaggar2021prottrans} for residue-level embeddings, followed by gated attention pooling and homology-based label diffusion using DIAMOND. We adapt this model for DNA sequence–to–phenotype prediction by replacing the protein encoder with GENERator~\citep{wu2025generator}, which produces residue-level embeddings from DNA sequences. These embeddings are pooled using the original gated attention mechanism to obtain gene-level representations. For the label diffusion step, we substitute DIAMOND with BLASTN~\citep{camacho2009blast+} and compute pairwise DNA sequence identity scores, which are used to diffuse phenotype labels across similar genes. The model is then trained to predict phenotype terms instead of GO functions.

\subsection{$*$InterLabelGO}
InterLabelGO+ is a hybrid approach that combines deep learning with alignment-based methods for protein function prediction, achieving strong performance in the CAFA5 challenge through its novel loss function that addresses label dependency and imbalance. We adapt this framework for DNA phenotype prediction by: (1) replacing protein embeddings with DNA sequence embeddings generated by GENERator \citep{wu2025generator}, (2) substituting DIAMOND \citep{buchfink2021diamond} with BLASTN \citep{camacho2009blast+} to compute pairwise DNA sequence identity scores and calculate weighted training set labels for test sequences, (3) retaining the effective ZLPR loss function which is equally suitable for our multi-label phenotype prediction task, and (4) replacing GO function terms with phenotype terms as target labels.

\section{Experiment details}
\label{app:experiment_details}
All experiments were conducted on two NVIDIA A6000 GPUs (48 GB each) running Ubuntu 22.04. Our codebase relies on PyTorch 2.4.1 and PyTorch Lightning 2.5.1; the complete list of dependencies is provided in \texttt{requirements.txt} (included in the code release upon acceptance).

For human datasets (GWAS, CAFA2, HPO), the average training time per epoch was approximately 2 minutes. On the MPO dataset, epoch duration increased to roughly 40 minutes due to the longer gene sequences in the mouse genome. We trained each model for 100 epochs and selected the checkpoint with the highest validation $F_{\max}$ for final evaluation.

\paragraph{Missing Input Handling}

As mentioned in Section~\ref{sec:introduction}, many existing phenotype-prediction methods rely on highly curated data, which limits their application in real practice. For example, PPI-based approaches require gathering (or conducting wet-lab experiment) every new sequence’s interactions with other $n$ proteins, which is an \(\mathcal{O}(n)\) complexity process that becomes infeasible as \(n\) grows.

In contrast, our sequence-based model needs no additional data collection or experiments for unseen sequences. When a gene lacks any GO annotations, we replace its fine-grain GO embedding with a random vector
\[
  \mathbf{h}_i \;\in\;\mathbb{R}^{1\times d},\quad d=1024,
\]
where \(d\) is the hidden dimension of GoBERT. In the bottleneck layer, we simply set that sample’s coarse-grain GO classification loss to zero:
\[
  \mathcal{L}_{\text{MLC}}^{\text{GO}} = 0.
\]
Empirically, even with approximately 20\% of genes missing GO annotations in the GWAS dataset, our model still outperforms most baselines on \(F_{\max}\) and AUROC, demonstrating the effectiveness of the complementary GO modality.  

\begin{table*}[!t]
\centering
\begin{tabular}{lccccccc}
\hline
 Dataset     & $\lambda_1$ & $\lambda_2$ & $\tau$ & batch\_size & hidden\_dim & learning\_rate & dropout \\ \hline
MPO   & [0.1-2]                         & [0.1-2]                         & 1                   & [12,16,20,24]          & [384, 512, 768,1024]         & [1e-6, 3e-6, 5e-6, 1e-5, 3e-5]           & [0.1-0.5]     \\ 
HPO   & [0.1-2]                       & [0.1-2]                       & 1                   & [12,16,20,24]          & [384, 512, 768,1024]         & [1e-6, 3e-6, 5e-6, 1e-5, 3e-5]           & [0.1-0.5]     \\ 
GWAS  & [0.1-2]                         & [0.1-2]                       & 1                   & [12,16,20,24]          & [384, 512, 768,1024]         & [1e-6, 3e-6, 5e-6, 1e-5, 3e-5]           & [0.1-0.5]     \\ 
CAFA2 & [0.1-2]                       & [0.1-2]                       & 1                   & [12,16,20,24]          & [384, 512, 768,1024]        & [1e-6, 3e-6, 5e-6, 1e-5, 3e-5]           & [0.1-0.5]     \\
\hline
\end{tabular}
\caption{The hyperparameter selection range during training stage. [low-high] stands for hyperparameter are ranged from low to high with 0.1 step size.}
\label{tab:hyper_selection}
\end{table*}

\begin{table*}[!t]
\centering
\begin{tabular}{lccccccc}
\hline
 Dataset     & $\lambda_1$ & $\lambda_2$ & $\tau$ & batch\_size & hidden\_dim & learning\_rate & dropout \\ \hline
MPO   & 1                         & 1                         & 1                   & 12          & 768         & 1e-5           & 0.1     \\ 
HPO   & 0.5                       & 0.5                       & 1                   & 16          & 768         & 3e-5           & 0.3     \\ 
GWAS  & 1                         & 0.1                       & 1                   & 16          & 512         & 1e-6           & 0.1     \\ 
CAFA2 & 0.4                       & 0.4                       & 1                   & 18          & 1024        & 5e-6           & 0.2     \\ 
\hline
\end{tabular}
\caption{The finalized hyperparameter on all datasets. Note the threshold for the $F_\text{max}$ is calculated following description in~\Cref{app:evaluation_metrics}.}
\label{tab:hyper_final}
\end{table*}

\section{Case studies: phenotype formation mechanisms from bottleneck weights}
\label{app:case_study}
To evaluate the interpretability of GenePheno and explore the biological mechanisms linking genetic information to phenotype formation, we conduct comprehensive case studies on the HPO, MPO, and GWAS datasets. For each dataset, we analyze the learned weights from the bottleneck layer, which captures general gene functions relevant to phenotypic outcomes. We examine the biological alignment between these functions and the corresponding phenotypes. As shown in~\Cref{fig:case_study}, we visualize selected examples of the bottleneck weights, highlighting $4\times4$ pairs of coarse-grained GO functions and phenotypes per dataset. These cases demonstrate how our model captures biologically meaningful mechanisms that connect genetic information to observable phenotypes.

\paragraph{MPO dataset}
The MPO dataset used in our study is derived from mouse gene knockout experiments, where phenotypic consequences are observed following targeted disruption of specific genes~\citep{smith2009mammalian}. GenePheno reveals several biologically plausible associations in the MPO dataset. Two representative cases shown in~\Cref{fig:case_study} (a) are described below. First, `maintenance of location” (GO:0051235) is associated with  abnormal aorta tunica media morphology” (MP:0009873). Abnormal aortic tunica media morphology is a common feature of diseases such as aortic aneurysms, characterized by disorganization of cellular and extracellular matrix structures, which reflects a failure in maintaining proper cell positioning~\citep{tang2005hyperplastic}. This supports the observed association between maintenance of location” and the abnormal aorta tunica media morphology”. Second, the interpretability results of GenePheno show that disruptions in isomerase activity” (GO:0016853) may contribute to increased energy expenditure” (MP:0004889), which is consistent with biological knowledge. For instance, prolyl isomerases such as Pin1 have been shown to regulate skeletal muscle metabolism by modulating SERCA activity, thereby influencing systemic energy use and exercise capacity~\citep{nakatsu2024prolyl}.

\paragraph{HPO dataset}
The HPO dataset used in our study is sourced from the official Human Phenotype Ontology database, where gene–phenotype associations are curated from medical publications and cover diverse aspects of human phenotypes~\citep{kohler2021human}. ~\Cref{fig:case_study}(b) shows sample bottleneck weights from the HPO dataset. We observe a strong association between the `molecular function inhibitor" (GO:0140678) and the decreased circulating complement C4 concentration" (HP:0045042). This is consistent with prior studies, which show that complement inhibitors can suppress C4 activity and reduce circulating levels~\citep{coss2023complement, garred2021therapeutic}. Another biologically plausible link is observed between postsynapse" (GO:0098794) and axial muscle weakness" (HP:0003327). Disruption of signal transmission at the neuromuscular junction (NMJ), particularly at the postsynaptic membrane, can impair muscle function and cause weakness~\citep{jones2017cellular, hirsch2007neuromuscular}.

\paragraph{GWAS dataset}
The GWAS dataset links genes to phenotypes through SNP associations~\citep{buniello2019nhgri, macarthur2017new}. Unlike HPO or MPO, which broadly capture genetic influences on diverse traits, GWAS data focus on a limited set of disease-related phenotypes. The weights for GenePheno interpretation are shown in~\Cref{fig:case_study}(c). The `growth cone membrane” (GO:0032580) refers to a specialized structure at the tip of extending axons that enables neurons to navigate their environment during development by responding to extracellular guidance cues~\citep{vitriol2012growth,igarashi2019molecular}. In the GWAS dataset, GenePheno reveals strong associations between this function and two neurologically related phenotypes: compulsive behaviors” (HP:0000722) and glioblastoma multiforme” (HP:0012174). GenePheno also highlights a strong association between cytokine production” (GO:0001816) and inflammation of the large intestine” (HP:0002037). Cytokines are key signaling molecules released during inflammatory responses that orchestrate immune cell recruitment and activate downstream pathways~\citep{zhang2007cytokines, sanchez2008role, escalante2025leaky}.

\section{Reproducibitily Details}

\label{app:reproducibility}
To better facilitate reproducibility, we provide all tunable model parameters, experimental settings, and our hyperparameter selection criteria.

\subsection{Random Seeds}
We initialize all random number generators using the PyTorch Lightning function ``lightning.pytorch.seed\_everything(seed=605)", which seeds PyTorch, NumPy, and Python’s \texttt{random} module.

\subsection{Range of Candidate Hyperparameters}
Table~\ref{tab:hyper_selection} lists each tunable hyperparameter and its candidate range. We select the optimal configuration based on the highest validation $F_{\max}$ for each dataset.

\subsection{Final Hyperparameters}

The selected hyperparameter values for each dataset are reported in Table~\ref{tab:hyper_final}.

\end{document}